\documentclass[sigconf,authorversion,nonacm]{acmart}

\newif\ifapx
\apxtrue %

\newcommand{\ourmaintitle}{\textsc{Seqret}: Mining Rule Sets from Event Sequences}

\newcommand{\ourmethod}{\textsc{Seqret}\xspace}
\newcommand{\oururl}{\url{https://eda.rg.cispa.io/prj/seqret/}}

\makeatletter
\newif\if@restonecol
\makeatother

\usepackage{latexsym}
\usepackage[ruled, vlined, nofillcomment, linesnumbered]{algorithm2e}
\usepackage[notheorems]{eda} %
\usepackage{prettyplots}

\usepackage{amsmath}
\usepackage{bm}
\usepackage{pgfplots}
\usetikzlibrary{calc}
\usetikzlibrary{backgrounds}
\usepgfplotslibrary{statistics}
\usepackage{pgfplotstable}
\usepackage{subcaption}
\usepackage{bbm}
\usepackage{soul}

\SetKwComment{tcpas}{\{}{\}}
\SetCommentSty{textnormal}
\SetArgSty{textnormal}
\SetKwRepeat{Do}{do}{while}
\SetKw{False}{false}
\SetKw{True}{true}
\SetKw{Null}{null}
\SetKwInOut{Output}{output}
\SetKwInOut{Input}{input}
\SetKw{AND}{and}
\SetKw{OR}{or}
\SetKw{Continue}{continue}
\SetKwIF{If}{ElseIf}{Else}{if}{:}{else if}{else}{end if}

\DeclareMathOperator*{\argmax}{arg\,max}

\makeatletter
\newcommand{\strequal}[2]{\pdf@strcmp{#1}{#2}==0}
\makeatother

\pgfplotsset{
	my legend/.style={
		mark=square*,
		fill,
		mark options={scale=4, fill opacity=0.5}
	}
}

\newcommand{\LN}{L_{\mathbb{N}}}

\newcommand{\set}[1]{\left\{#1\right\}}

\newcommand{\ourmethodcandidates}{\textsc{Seqret-Candidates}\xspace}
\newcommand{\ourmethodmine}{\textsc{Seqret-Mine}\xspace}
\newcommand{\methodname}[1]{\textsc{#1}\xspace}
\newcommand{\varname}[1]{\ensuremath{\mathit{#1}}\xspace}

\newcommand{\rheadd}[1]{\ensuremath{\mathit{head}(#1)}\xspace}
\newcommand{\rtaill}[1]{\ensuremath{\mathit{tail}(#1)}\xspace}
\newcommand{\supp}[1]{\ensuremath{\mathit{supp}(#1)}\xspace}
\newcommand{\conf}[1]{\ensuremath{\mathit{conf}(#1)\xspace}}
\newcommand{\triggercount}[1]{\ensuremath{\mathit{trigs}(#1)}\xspace}

\newcommand{\patternsimilarity}[2]{\ensuremath{\mathit{sim}\left(#1,#2\right)}\xspace}
\newcommand{\rulesimilarity}[2]{\ensuremath{\mathit{sim}\left(#1,#2\right)}\xspace}
\newcommand{\LCSdistance}[2]{\ensuremath{\mathit{d}\left(#1,#2\right)}\xspace}
\newcommand{\recall}[2]{\ensuremath{\mathit{recall}\left(#1,#2\right)}\xspace}
\newcommand{\precision}[2]{\ensuremath{\mathit{precision}\left(#1,#2\right)}\xspace}

\newcommand{\jmlr}{\textit{\usefont{OT1}{cmr}{m}{it}JMLR}\xspace}

\newcommand{\maxgap}{\emph{max gap}\xspace}
\newcommand{\maxdelay}{\emph{max delay}\xspace}
\newcommand{\windoworder}{\methodname{window order}}
\newcommand{\pruneorder}{\methodname{prune order}}
\newcommand{\searchorder}{\methodname{extend order}}

\newcommand{\bestrwins}{\methodname{BestRuleWin}}
\newcommand{\nextrwin}{\methodname{NextBestWin}}
\newcommand{\cover}{\methodname{Cover}}
\newcommand{\miner}{\methodname{Seqret}}
\newcommand{\prune}{\methodname{Prune}}
\newcommand{\signineigh}{\methodname{CandRules}}
\newcommand{\signineighshort}{\methodname{CandRules}}

\newcommand{\gapi}[2]{g(#1,#2)}

\newcommand{\out}[1]{\st{#1}}

\newcommand{\sqssqrt}{\textsc{Seqret-Candidates}\xspace}
\newcommand{\tns}{\textsc{TNS}\xspace}
\newcommand{\sqs}{\textsc{Sqs}\xspace}
\newcommand{\poerma}{\textsc{Poerma}\xspace}
\newcommand{\poermh}{\textsc{Poermh}\xspace}
\newcommand{\cossu}{\textsc{Cossu}\xspace}

\begin{document}

\title{\ourmaintitle}

\author{Aleena Siji}
\authornote{Work done while at CISPA Helmholtz Center for Information Security.}
\email{aleena.siji@helmholtz-munich.de}
\affiliation{%
	\institution{Helmholtz AI}
	\country{Germany}
}

\author{Joscha C\"{u}ppers}
\email{joscha.cueppers@cispa.de}
\affiliation{%
	\institution{CISPA Helmholtz Center for Information Security}
	\country{Germany}
}

\author{Osman Ali Mian}
\authornotemark[1]
\email{osman.mian@uk-essen.de}
\affiliation{%
\institution{Institute for AI in medicine IKIM}
	\country{Germany}
}

\author{Jilles Vreeken}
\email{jv@cispa.de}
\affiliation{%
	\institution{CISPA Helmholtz Center for Information Security}
	\country{Germany}
}

\begin{abstract}
  Summarizing event sequences is a key aspect of data mining. Most existing methods neglect conditional dependencies and focus on discovering sequential patterns only. In this paper, we study the problem of discovering both conditional and unconditional dependencies from event sequence data. We do so by discovering rules of the form $X \rightarrow Y$ where $X$ and $Y$ are sequential patterns. Rules like these are simple to understand and provide a clear description of the relation between the antecedent and the consequent. To discover succinct and non-redundant sets of rules we formalize the problem in terms of the Minimum Description Length principle. As the search space is enormous and does not exhibit helpful structure, we propose the \ourmethod method to discover high-quality rule sets in practice. Through extensive empirical evaluation we show that unlike the state of the art, \ourmethod ably recovers the ground truth on synthetic datasets and finds useful rules from real datasets.

\end{abstract}

\maketitle

\section{Introduction}
\label{sec:intro}

\renewcommand{\out}[1]{}

In many applications data naturally takes the form of events happening over time. Examples include industrial production logs, the financial market, device failures in a network, etc. Existing methods for analyzing event sequences primarily focus on mining unconditional, frequent sequential patterns~\cite{agrawal:95:mining,mannila:00:global,tatti:09:signifeps}. Loosely speaking, these are subsequences that appear more often in the data than we would expect. Real world processes are often more complex than this, as they often include conditional dependencies. The formation of tropical cyclones ($C$) in the Bay of Bengal, for example, is often but not always followed by heavy rainfall ($R$) on the coast. Knowing such a relationship is helpful both in predicting events and in understanding the underlying data generating mechanisms.

In this paper, we are interested in discovering rules of the form $X \rightarrow Y$ from long event sequences, where $X$ and $Y$ are sequential patterns. Existing methods for mining such rules either suffer from the pattern explosion, i.e. are prone to returning orders of magnitude more results than we can possibly analyze~\cite{poerm,poermh}, or are strongly limited in the expressivity, e.g. require the constituent events to occur in a contiguous order~\cite{cossu}. 

We aim to discover succinct sets of rules that generalize the data well. We explicitly allow for gaps between the head and the tail of the rule, as well as in the occurrences of $X$ and $Y$ themselves. To ensure we obtain compact and non-redundant results, we formalize the problem using the Minimum Description Length (MDL) principle~\cite{grunwald:07:book}. Loosely speaking, we are after that set of sequential rules that together compresses the data best. 

However, the problem we so arrive at is computationally challenging. For starters, there exist exponentially many rules, exponentially many rule sets, and then again exponentially many ways to describe the data given a set of rules. Moreover, the search space does not exhibit structure we can use to efficiently obtain the optimal result. To mine good rule sets from data we therefore propose the greedy \ourmethod algorithm. We introduce two variants. \ourmethodcandidates constructs a good rule set from a set of candidate patterns by splitting them into high-quality rules. \ourmethodmine, on the other hand, only requires the data and mines a good rule set from scratch. Starting from a model of singleton rules, it iteratively extends them into more refined rules. To avoid testing all possible extensions, we consider only those extensions that occur significantly more often than expected. 

Through extensive evaluation, we show that both variants of \ourmethod work well in practice. On synthetic data we show that they are robust to noise and recover the ground truth well. On real-world data, we show that \ourmethod returns succinct sets of rules that give clear insight into the data generating process. This in stark contrast to existing methods which either return many thousands of rules \cite{poerm} or are restricted to rules where events occur contiguously \cite{cossu}.

To summarize, the main contributions are as follows:
\begin{enumerate}[noitemsep,topsep=0pt,label=(\alph*)]
    \item We define a pattern language for fully ordered sequential rules that accommodates for gaps.
    \item We present \ourmethodcandidates for constructing a high-quality rule set from a given set of sequential patterns.
    \item We present \ourmethodmine for mining a high-quality rule set given a databse of event sequences. 
    \item We extensively evaluate \ourmethod on synthetic and real-world datasets, comparing it to the state-of-the-art.
\end{enumerate}

We make all code and data available online.\!\footnote{\href{https://eda.rg.cispa.io/prj/seqret/}{https://eda.rg.cispa.io/prj/seqret/}} %

\section{Preliminaries}
\label{sec:prelim}

In this section we introduce basic notation and give a short introduction to the MDL principle.

\subsection{Notation}
As data we consider a sequence database $D$ of $|D|$ event sequences. A sequence $S \in D$ consists of $|S|$ events drawn from a finite alphabet \(\Omega\) of discrete events $e \in \Omega$. We denote the total number of events in the data as $||D||$. 
We write $S_t$ for the $t^\mathit{th}$ sequence in $D$. To avoid clutter, we omit the subscript whenever clear from context. We write $S[i]$ to refer to the $i^\mathit{th}$ event in sequence $S$, and $S[i,j]$ for the subsequence from the $i^\mathit{th}$ up to and including the $j^\mathit{th}$ event of $S$. We denote an empty sequence by $\epsilon$.

A serial episode $X$ is a sequence of $|X|$ events drawn from $\Omega$. 
A sequential rule $r$ captures the conditional dependence between a serial episode $X$ and a serial episode $Y$. Intuitively, it expresses that whenever we see $X$ in the data it is more likely that $Y$ will follow soon.
We refer to \(X\) as the \emph{head} or antecedent of $r$, denoted \rheadd{r}, and to $Y$ as the \emph{tail} or consequent of $r$, denoted \rtaill{r}. If $X$ is an empty pattern, $X = \epsilon$, we call $X \rightarrow Y$ an \emph{empty head rule}. We refer to empty head rules where $|Y| = 1$ as a \emph{singleton} rule. 

A subsequence $S[i,j]$ is a window of pattern $X$ iff $X$ is a subsequence of $S[i,j]$, and subsequently we say $S[i,j]$ \emph{matches} $X$ and vice-versa we say that $X$ occurs in $S[i,j]$. A pattern window \(S[i,j]\) is \emph{minimal} for \(X\) iff no proper sub-window of \(S[i,j]\) matches \(X\). A window of a rule $r$ is a tuple of two pattern windows $S[i,j]$ and $S[k,l]$ when $S[i,j]$ matches $\rheadd{r}$, $j < k$, and $S[k,l]$ matches $\rtaill{r}$. We denote a rule window by $S[i,j;k,l]$.

\begin{figure}[t]
	\centering
	\includegraphics[width=0.45\textwidth]{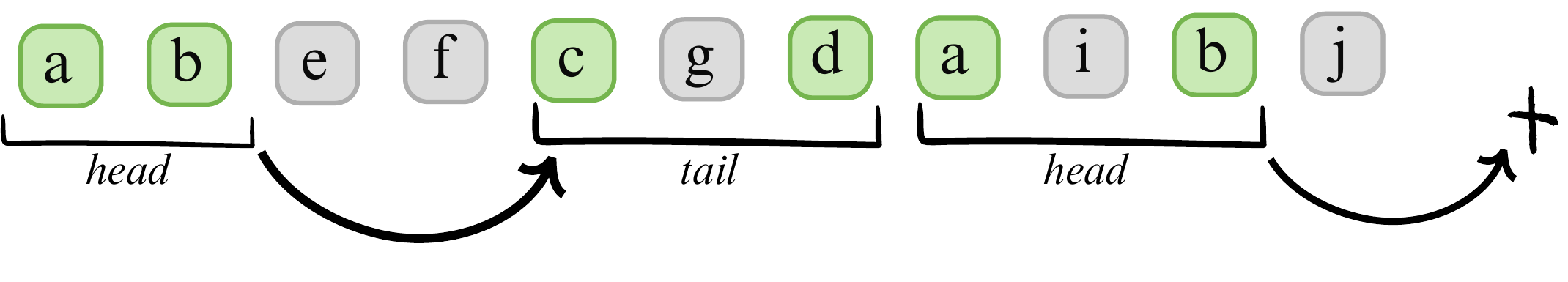}
	\caption{Toy example of a rule \(ab \rightarrow cd\) in an event sequence. Each occurrence of head \(ab\) triggers the rule. The first is followed by tail \(cd\) and hence a `hit' whereas the second is not and hence a `miss'. $\supp{ab \rightarrow cd} = 1$ and $\conf{ab \rightarrow cd} = 0.5$. %
	}
	\label{fig:examplerule}
\end{figure}

We say a window $S[i,j]$ \emph{triggers} rule $r$ when it is a minimal window of $\rheadd{r}$. A rule window $S[i,j;k,l]$ \emph{supports} a rule $r$ if $S[i,j]$ triggers \rheadd{r} and $S[k,l]$ matches \rtaill{r}. We call the number of events that occur in a rule window between the rule head and the rule tail, $k-j-1$, the \emph{delay} of the rule instance. We give an example in Fig.~\ref{fig:examplerule}.
We denote the number of windows over all sequences $S \in D$ that trigger a rule $r$ as the trigger count \triggercount{r}. 
We define the \emph{support} of a rule $r$ as the number of rule windows $S[i,j;k,l]$ in $D$ where $S[k,l]$ is a minimal window of $\rtaill{r}$ and follows the head with minimum delay. Finally, we define the confidence of a rule $r$ as its support relative to its trigger count, formally $\conf{r} = \supp{r}/\triggercount{r}$.

\subsection{Minimum Description Length Principle} 
The Minimum Description Length (MDL) \cite{grunwald:07:book,rissanen} is a computable and statistically well-founded version of Kolmogorov complexity \cite{kol1}. For a given model class $\mathcal{M}$ it identifies the best model $M \in \mathcal{M}$ as the one that minimizing the number of bits for describing both model and data without loss, or formally, $L(M) + L(D \mid M)$ with $L(M)$ the length in bits of model $M$ and $L(D\mid M)$ the length in bits of data $D$ given $M$. 
This is known as two-part, or crude MDL---in contrast to one-part, or refined MDL~\cite{grunwald:07:book}, which is not computable for arbitrary model classes. We use two-part MDL because we are interested in the model: the set of rules that describe the data best. In MDL we are never concerned with materialized codes, we only care about code lengths. To use MDL, we have to define a model class and code length functions for models and data given a model.

\section{MDL for Sequential Rules}
\label{sec:theory}
We now formally define the problem we aim to solve. We consider sets $R$ of sequential rules as our model class $\mathcal{R}$. By MDL, we are interested in that set of rules $R \in \mathcal{R}$ that best describes data $D$. 

\begin{figure}
	\centering
	\includegraphics[width=0.8\linewidth]{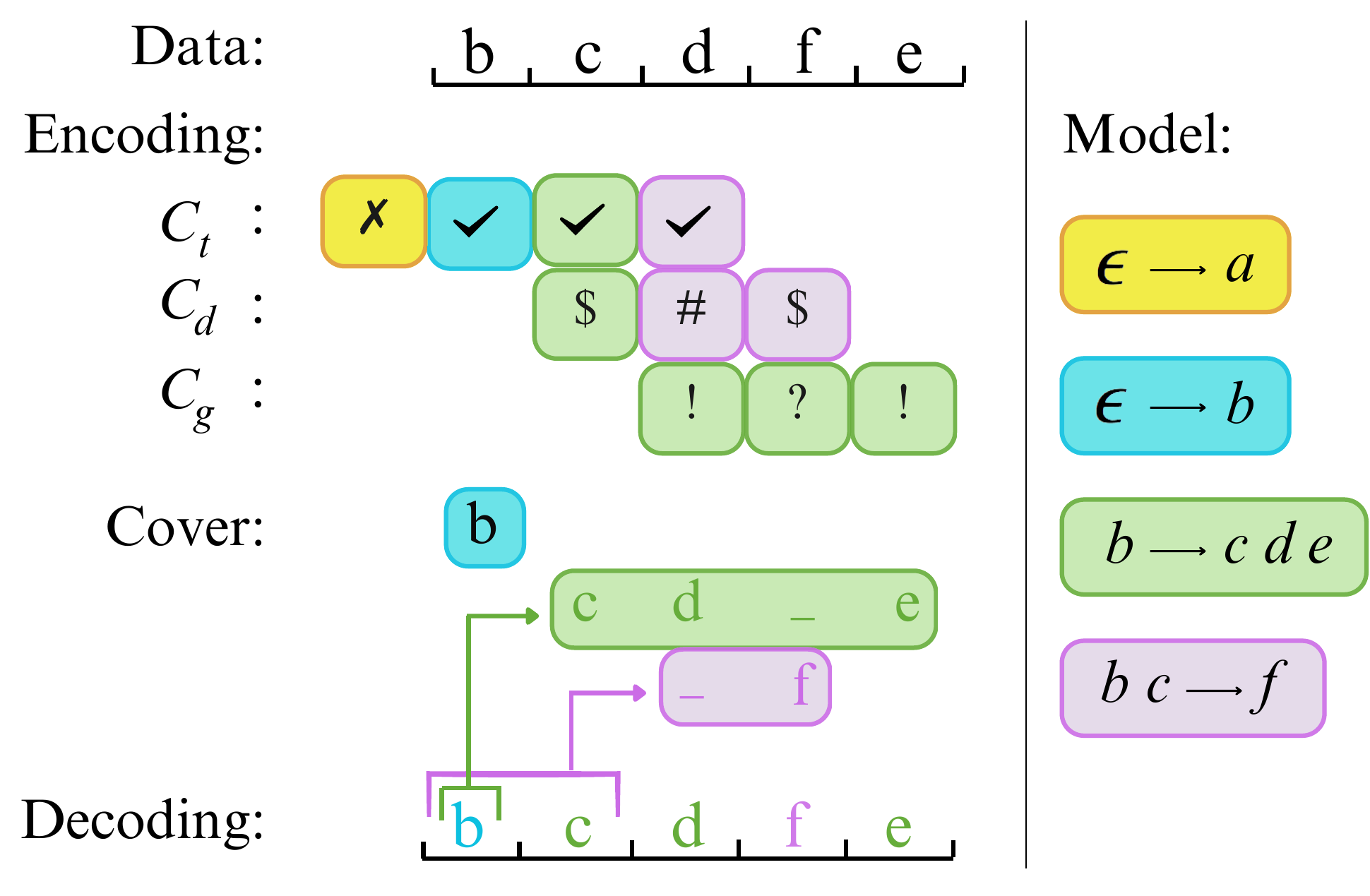}
	\caption{Toy example showing an encoding of sequence $S$ using rule set $R$. The encoding consists of three code streams. $C_t$ encodes if a triggered rule \emph{hits} or \emph{misses}. $C_d$ encodes the delay between the trigger and the rule tail. $C_g$ encodes the gaps in the tails. Together, they form a cover $C$ of $D$ given $R$.}
	\label{fig:encoding-decoding}
\end{figure}

\subsection{Decoding an Event Sequence}
Before we formally define how we \emph{encode} models and data, we give the main intuition of our score by \emph{decoding} an already encoded sequence. We give an example in Fig.~\ref{fig:encoding-decoding}.

To decode a symbol, we consider the rules from $R$ that are currently triggered. For those, we read codes from the trigger stream $C_t$. Initially, the context is empty and hence only empty-head rules trigger. The first trigger code is a \emph{miss} for singleton rule $\epsilon \rightarrow a$. The second trigger code is a \emph{hit} for rule $\epsilon \rightarrow b$. As empty-head rules do not incur delays, we can write $b$ as the first symbol of the sequence. 

This triggers rule $b \rightarrow \mathit{cde}$. We hence read a code from the trigger stream, and find that it is a hit. As this rule does not have an empty head, there may be a delay between the head and its tail and we read a code from the delay steam $C_d$ to determine if this is the case. It is a start code, so we write the first symbol of the tail ($c$). 

This creates a minimal window of \(bc\) and hence rule \(bc \rightarrow f\) triggers. We read from $C_t$ to find that it hits, and from $C_d$ to find that its tail is delayed. To determine if we may write the next symbol from tail $cde$, we read from the gap stream $C_g$. This is a fill code, meaning there is no gap, and hence we write $d$. 

This time, no new rule triggers. Tail $cde$ is not yet completely decoded and $f$ is delayed. For each delayed tail we read a code from $C_d$, and for each incomplete tail we read a code from $C_g$. Here, we read a start code for tail $f$ and a gap code for tail $cde$, we hence write $f$. Again, no new rule is triggered. Now only tail $cde$ is not yet fully decoded. We read from $C_g$ and as it is a fill code we write $e$ as the last symbol of the sequence.

To summarize, sequences are encoded from left to right, and rules automatically trigger whenever we observe a minimal window of the head. For each trigger, we encode whether the tail follows using a \emph{hit} or \emph{miss} code. When a rule hits, we encode whether its tail follows immediately or later, using a \emph{start} resp. \emph{delay} code. Finally, we encode whether \emph{gaps} occur in the rule tail using \emph{fill} and \emph{gap} codes. Empty-head rules never incur a delay. To avoid unnecessary triggers, we only encode those of empty-head rules if no other rule encodes the current symbol (e.g. all active tails say `gap').

\subsection{Computing the Description Lengths}
Now that we have the intuition, we can formally describe how to encode a model, respectively the data given a model. 

\paragraph{Encoding a Model}
A model $R \in \mathcal{R}$ is a set of rules. To reward structure between rules, e.g. chains where the tail of one rule is the head of another (e.g. $r_1 = \epsilon \rightarrow AB$, and $r_2 = AB \rightarrow CD$), we first encode the set $P$ of all non-empty and non-singleton heads and all non-singleton tails. Formally, 
\[
P = \set{\rheadd{r} \mid \forall r \in R} \cup \set{\rtaill{r} \mid \forall r \in R} \setminus (\Omega \cup \epsilon) \; .
\]
The encoded length $L(P)$, is defined as
\[
L(P) = \LN( |P| + 1) + \sum_{p \in P} \LN(|p|) + |p| \log_2(|\Omega|)\;, 
\] 
where we first encode the number of these patterns using $\LN$ the MDL-optimal encoding for integers \cite{rissanen:83:integers}. It is defined for $z \geq 1$ as $\LN(z) = \log^* z + \log c_0 \,$ where $\log^*z $ is the expansion $\log z + \log \log z + \cdots$ where we only include the positive terms. To ensure this is valid encoding, i.e. one that satisfies the Kraft inequality, we set $c_0 = 2.865064$~\cite{rissanen:83:integers}. Since $P$ can be empty and $\LN$ is only defined for numbers $\geq 1$ we offset it by one. 
Next, we encode each pattern $p \in P$ where we use $\LN$ to encode its length and then choose each subsequent symbol $e \in p$ out of alphabet $\Omega$. 

Now that we have the set of all heads and tails, we have 
\[
L(R \mid P) = \LN(\vert R \vert + 1) + |R| (\log_2(\vert P\vert + \vert \Omega\vert + 1) + \log_2(\vert P\vert + \vert \Omega\vert)) \;,
\]
as the encoded length in bits of a set of rules. We first encode the number of rules, and as $R$ can be empty, we again offset by one. Next, for each rule $r \in R$, we choose its head from $P \cup \Omega$, and then its tail from $P \cup \Omega$. 

Putting this together, the number of bits to describe a rule set $R \in \mathcal{R}$ without loss is
\[
L(R) = L(P) + L(R \mid P) \;.
\]

\paragraph{Encoding Data given a Model}
As we saw in the example, to reconstruct the data we need the three code streams $C_t$, $C_d$, and $C_g$. For an arbitrary database we additionally need to know how many sequences it includes, and how long these are. Formally, the description length of data $D$ given a model $R$ hence is
\begin{equation}
\label{eq:totaldata}
    \begin{aligned}
        L(D \mid R) =  \LN(|D|) + \left( \sum_{S \in D} \LN(|S|) \right) + L(C_t) + L(C_d) + L(C_g) \; .
    \end{aligned}
\end{equation}

To encode the code streams $C_t, C_d, C_g$ we use prequential codes~\cite{grunwald:07:book}. These codes are asymptotically optimal without requiring us to make arbitrary choices in how to encode the code distributions. Formally, we have 
\[
L(C_j) = \sum_{i = 1}^{|C_j|} \log_2 \frac{\mathit{usg}_i(C_j[i] \mid C_j) + c}{i + \mathit{unique}(C_j) \cdot c} \quad,
\]
where $\mathit{usg_i}(C_j[i]\mid C_j)$ denotes the number of times $C_j[i]$ has been used in $C_j$ up to the $i^{\mathit{th}}$ position, $\mathit{unique}(C_j)$ denotes the number of unique symbols in $C_j$, and $c$ is a small constant. As is common in prequential coding, we set $c$ to $0.5$.

\subsection{The Problem, Formally}
We can now formalize the problem we aim to solve.

\vspace{0.5em}
\noindent\textbf{The Sequential Rule Set Mining Problem}
\emph{
    Given a sequence database \(D\) over alphabet $\Omega$, find the smallest rule set \(R \in \mathcal{R}\) and cover $C$ such that the total encoded size 
    \[L(R) + L(D \mid R)\] 
    is minimal.
	}
\vspace{0.5em}

The search space of this problem is enormous. To begin with, there exist super-exponentially many covers of $D$ given $R$. The optimal cover depends on the code lengths, which in turn depend on the code usages. Even if the optimal cover is given, the problem of finding the optimal rule set is super-exponential: there exist exponentially many patterns $p$ in the size of the alphabet $\Omega$, exponentially many rules $r$ in the number of patterns, and exponentially many sets of rules. None of these sub-problems exhibit substructure, e.g. monotonicity or submodularity, that we can exploit to efficiently find the optimal solution. Hence, we resort to heuristics.

\section{The Seqret Algorithm}
\label{sec:algo}

In this section we introduce our method, \ourmethod, for discovering high-quality \textbf{seq}uential \textbf{r}ule-s\textbf{et}s from data. 
We break the problem down into two parts: optimizing the description of the data given a rule set, and mining good rule sets. For the latter we propose \ourmethodcandidates for doing so given a set of candidate patterns, and \ourmethodmine for mining rule sets directly from data.

\subsection{Selecting a Good Cover}

A lossless description of $D$ using rules $R$ correspond to a set of rule windows such that each event $e$ in $D$ is covered by exactly one window. We are after that \emph{cover} that minimizes $L(D \mid R)$. Finding the optimal cover is infeasible, and hence we instead settle for a good cover and show how to find one greedily.

The main idea is to define an order over the rule windows and greedily select the next best window until the data is completely covered. To minimize the encoded length, we prefer to cover as many events as possible with a single rule with few gaps. Therefore, we prefer using rules  with long tails, high confidence, and high support. Similarly, among windows of otherwise equally good rules, we prefer those with lower delays and fewer gaps in the rule tail. As a final tie breaker, we consider the starting position of the rule tail. Combining this, we define the \windoworder as descending on $|\rtaill{r}|$, $\conf{r}$, and $\supp{r}$, and finally ascending on $l-j-|\rtaill{r}|$, and $k$, where $r$ is a rule, $S[i,j;k,l]$ is a rule window. To avoid searching for all possible rule windows, we start with the best window per rule trigger and look for the next best only if we do not select the former due to conflicts, i.e. its constituent events are already covered by a previously selected window. We define the best rule window per trigger as the one with the fewest gaps in its rule tail window, and among those with same gap count, the one with the lowest delay. 

We give the pseudocode of \methodname{Cover} as Algorithm~\ref{alg:greedycover}. We start by initializing cover $C$ with the empty set and
window set $W$ with for each rule the best rule windows per trigger (lines 1-2). We then greedily add rule windows to $C$ in order of \windoworder. If a window conflicts with an already selected window (line \ref{alg:greedycover:conflict}), we skip it and search for the next best rule window for the corresponding trigger and add it to $W$ (line \ref{alg:greedycover:nextbest}). We continue this process until all events in $D$ are covered. To avoid evaluating hopeless windows, we limit ourselves to those within a user-set \maxdelay ratio and \maxgap ratio. We provide further details and pseudo code for \bestrwins and \nextrwin procedures in Appx.~\ref{apx:sec:rule-window}.

The worst case time complexity of \methodname{Cover} depends on the number of rules in $R$, total number of events in $D$, and the lengths of the heads and tails per rule. %
In Appx.~\ref{solcomplexity} we show the complexity of \methodname{Cover} is $\mathcal{O}(\vert R \vert \cdot||D||  (h+ t^3 + t  \log_2(\vert R \vert \cdot ||D|| t))$, where $h$ is the max head length and $t$ the max tail length. 

Next, we consider the problem of discovering good rule sets. 

\begin{algorithm}[t]
\caption{\methodname{Cover}}
\label{alg:greedycover}
\SetAlgoNoEnd
\SetAlgoLined
\SetCommentSty{mycommfont}
\SetKwRepeat{Do}{do}{while}
\KwIn{Sequence database $D$, rule set $R$}
\KwOut{Cover $C$}
$C \gets \set{}$\; \label{alg:greedycover:c-init}
$W \gets \set{\bestrwins(r,D) \mid \forall r \in R}$\; \label{alg:greedycover:w-init}
\While{$\exists S_t \in D \text{ where, } \exists e \in S_t \text{ not covered by } C$}{ \label{alg:greedycover:while}
    $w \gets$ next $w \in W$ in \windoworder;\\
    $W \gets W \setminus \{w\}$;\\ %
    \If{$\nexists z \in C$ that \textbf{conflicts} with $w$}{ \label{alg:greedycover:conflict}
        $C \gets C \cup \set{w}$\;
    }
    \Else{$W \gets W \cup \set{\nextrwin(w,C,D)}$\; \label{alg:greedycover:nextbest}
    }
}
\Return \(C\)
\end{algorithm}

\subsection{Selecting Good Rule Sets}
\begin{algorithm}[tb!]
    \caption{\ourmethodcandidates}
    \label{alg:candidates}
    \KwIn{Sequence database $D$, set of patterns $F$}
    \KwOut{Rule set $R$}
        $R \gets \set{\epsilon \rightarrow e \mid \forall e \in \Omega}$\\
        \For{$p \in F$ ordered descending by $L(D,F\setminus\{p\}) - L(D,F)$}{
            $r \leftarrow \argmax_{r' \in \textsc{Split}(p)} L(D,R) - L(D, R \cup \set{r'})$\\
            \If{$L(D, R \cup \set{r}) < L(D,R)$}{ 
                $R \gets M\cup \set{r}$\;
            }
        }
    \Return{$R$}\;
\end{algorithm}

We first propose an approach that does so given a set of sequential patterns as input. We start from the intuition that, if the ground truth includes a sequential rule $a \rightarrow bc$, a good sequential pattern miner will return $\mathit{abc}$. This means we can reconstruct ground truth rules by considering splits of candidate patterns $XY$ into candidate rules $X \rightarrow Y$ and using our score to select the best split.

To this end, we propose \ourmethodcandidates, for which we give the pseudocode as Algorithm \ref{alg:candidates}. 
We initialize the rule set with all singleton rules (line 1) to ensure we can encode the data without loss. 
We then iterate over each candidate pattern (line 2) in descending order of contribution to compression~\cite{sqs}.  %
We split each pattern into candidate rules -- for example, pattern \emph{abc} generates candidate rules $\epsilon \rightarrow \mathit{abc}$,$ a \rightarrow bc $, and $ ab \rightarrow c$ -- and choose the candidate rule that minimizes our score (line 3). We add it to our model if it improves the score (line 5) and iterate until all patterns are considered.

The run time is dominated by the number of cover computations, i.e. how many times we have to compute $L(D|R)$. We have to compute a new cover for each rule we test, and each pattern $p$ can be split into $|p|$ rules. We test each $p \in F$ as such the complexity of \ourmethodcandidates is $\mathcal{O}(|F|(\max_{p \in F}|p|))$.

\subsection{Generating Good Rules}
\label{solgreedyminer}
Next, we move our attention to generating good rule sets directly from data. The first step is to generate good candidate rules. Given a rule $r$ from the current model, we consider extending it with events $e \in \Omega$ that occur significantly more often within or directly adjacent to the rule windows of $r$. A rule has $|r|+1$ such gap positions, i.e. before its first event, between its constituent events, and after its last event. For example, rule $ab \rightarrow cde$ has 6 gap positions, 
\newcommand{\gap}[3]{
    \node at ($(#1.south east)!0.5!(#2.south west)$) [color=gray] {\rotatebox{-90}{\Huge$\prec$}};
    \node at ($(#1.south east)!0.5!(#2.south west)$) [below=3pt] {#3};
}

    \begin{center}
        \begin{tikzpicture}[
            scale=0.7
        ]
            \node[anchor=east] (start) at (0,0) {\strut};
            \node (a) at (1,0) {$a\strut$};
            \node (b) at (2,0) {$b\strut$};
            \node (r) at (2.75,0) {$\rightarrow$};
            \node (c) at (3.5,0) {$c\strut$};
            \node (d) at (4.5,0) {$d\strut$};
            \node (e) at (5.5,0) {$e\strut$};
            \node (end) at (6.5,0) {$.\strut$};
            \gap{start}{a}{$g_0$}
            \gap{a}{b}{$g_1$}
            \gap{b}{c}{$g_2$}
            \gap{c}{d}{$g_3$}
            \gap{d}{e}{$g_4$}
            \gap{e}{end}{$g_5$}
        \end{tikzpicture}
    \end{center}

For each rule $r$ we test for every gap position $g_i$ if $e \in \Omega$ is more frequent than expected in its rule windows. Our null hypothesis is
\[
H_0: \sum_{w \in B} \mathbbm{1}(e \in \gapi{i}{w}) \leq \sum_{w \in B} {\Pr}(e \in \gapi{i}{w})
\]
where $B$ is the set of best rule windows of $r$, $B = \bestrwins(r,D)$, $w \in B$ is a window of rule $r$, and $\gapi{i}{w}$ a function that returns gap $i$ from window $w$.

When computing the probability of an event $e$ in gap $g_i$, we have to account for differences in lengths of gaps between different windows. The probability of $e$ occurring in gap $g_i$ of a rule window $w_p$ is given by 
\[\Pr(e \in \gapi{i}{w_p}) = 1 - \left(1 - \frac{\supp{\epsilon \rightarrow e}}{|D|}\right)^{|\gapi{i}{w_p}|}\; .
\]
To test for statistical significance, we can model the expected neighborhood as a Poisson binomial distribution~\cite{pbdwang}. That is, the trials are the rule windows, and the success probability per trial is decided by the length of the gap at the position of interest. Computing the CDF of the Poisson binomial distribution is expensive~\cite{pbdapproxhong,lecamapprox,refinednormal}. As a fast approximation, we use the normal approximation with continuity correction \cite{pbdapproxhong} for cases where the number of trials, i.e \supp{r}, is greater than 10. 
If less than or equal to 10, we simply check if the actual count of occurrences is greater than the expected count by more than one. 

If event $e$ is measured to be significantly more frequent in $g_i$ than expected, we generate a new rule by inserting $e$ at the position of $g_i$ in the rule. We give the pseudo-code in Appx.~\ref{apx:sec:candidates}.

\subsection{Mining Good Rule Sets} 
Finally, we describe \ourmethodmine for mining good rule sets directly from data. We provide the pseudocode as Algorithm~\ref{alg:greedyminer}. We initialize rule set $R$ with all the singleton rules (line \ref{seqret:init}). Next, we consider adding candidates based on the rules already in the model (line \ref{seqret:extend}). As we want to generate the most promising candidate rules first, we start with rules with high support and high confidence. We define a greedy \searchorder as 1) $\uparrow \supp{r}$, 2) $\uparrow \conf{r}$, 3) $\uparrow |\rtaill{r}|$ and 4) $\uparrow |\rheadd{r}|$, where $\uparrow$ indicates that higher values are preferred. For each we generate a set of candidate rules as described above, we test them for addition in the order of their p-values~(line \ref{seqret:candidates}).

We add those rules into the model whose inclusion results in a significant reduction in the total encoded size (line \ref{seqret:test}). We use the no-hypercompression inequality~\cite{nohyper,grunwald:07:book} to test for significance at level $\alpha$, writing $\ll_\alpha$ for ``significantly less''.\!\footnote{In our experiments we set $\alpha$ to 0.05, which by the no-hypercompression inequality corresponds to a minimum gain of 5 bits.} 
In case adding a candidate rule $r'$ to the model does not improve compression, we test if replacing the rule $r$ we generated it from by $r'$ leads to a better compression (line \ref{seqret:replace-test} and \ref{seqret:replace}). To ensure we can always describe the data without loss, we never remove singleton rules. 

After adding a new rule, \ourmethodmine performs a pruning step to remove existing rules that may have become redundant or obsolete (line \ref{seqret:prune}). The \prune method iterates over the non-singleton rules in the model and removes those whose exclusion reduces the total encoded size. We do so in \pruneorder where we consider rules in order of lowest usage, highest encoded size, and lowest tail length.

We repeat generating candidate rules, adding them, and pruning redundant rules until convergence. Convergence is guaranteed as our score is lower bounded by 0.
The worst case time complexity of one iteration of \ourmethodmine is, $\mathcal{O}(\vert R \vert  \vert \Omega \vert (h + t))$. We provide the full derivation in Appx.~\ref{solcomplexity}.

\begin{algorithm}[h]
\caption{\ourmethodmine}
\label{alg:greedyminer}
\SetAlgoNoEnd
\SetAlgoLined
\SetCommentSty{mycommfont}
\SetKwRepeat{Do}{do}{while}
\KwIn{Sequence database \(D\) over \(\Omega\), significance level $\alpha$}
\KwOut{Rule set \(R\)}
$R \gets \set{\epsilon \rightarrow e \mid \forall e \in \Omega}$\; \label{seqret:init}
\Do{$R$ updated}{
    \For{$r \in R$ in \searchorder}{ \label{seqret:extend}
        \For{$r' \in \signineighshort(D,r)$ in order of $p$-value}{ \label{seqret:candidates}
            \If{$L(D,R\cup \set{r'}) \ll_\alpha L(D,R)$}{ \label{seqret:test}
                $R \gets R\cup \set{r'}$\;
            }
            \ElseIf{\(r \notin \set{\epsilon \rightarrow e \mid \forall e \in \Omega}\) \textbf{and} $L(D,R \cup \{r'\}\setminus \{r\}) \ll_\alpha L(D,R)$}{ \label{seqret:replace-test}
                $R \gets (R \setminus \set{r})\cup \set{r'}$\; \label{seqret:replace}
            }
            \If{$R$ updated}{
                $R \gets \prune(D,R)$\; \label{seqret:prune}
                \textbf{continue} with next $r$
            }
        }
    }
} \label{whilecondition}
\Return \(R\)
\end{algorithm}

\section{Related Work}
\label{sec:related}
Classical methods for mining sequential patterns focus on extracting all frequent patterns and therewith suffer from the well-known pattern explosion leading to excessively many, largely redundant, and often spurious results~\cite{surveyseq}. Mining closed frequent patterns alleviates this problem, but is sensitive to noise~\cite{yan2003clospan,wangBIDEEfficientMining2004}. Mining significant patterns \cite{low-kamMiningStatisticallySignificant2013,petitjeanSkopusMiningTopk2016,tononPermutationStrategiesMining2019,jenkinsSPEckMiningStatisticallysignificant2022} changes the objective to patterns of which the frequency is significant under a null hypothesis, usually that events occur independent from each other.

Pattern set mining avoids the pattern explosion by instead scoring \emph{sets} of patterns. The Minimum Description Length principle has been shown to be a robust criterion for identifying good patterns sets in practice~\cite{galbrun2022minimum, cuppers2024flowchronicle, sqs}.
For event sequences, different pattern languages, scores, and methods have been proposed. There exist those that allow for gaps \cite{sqs}, interleaved patterns \cite{squish,hopper}, generalized patterns \cite{flock}, and periodic patterns \cite{galbrunMiningPeriodicPatterns2018}. 

Classical rule miners for event sequences operate similar to frequent pattern mining, but in addition to the frequency requirement also impose a minimum confidence threshold. Various approaches have been proposed to address different data modalities, such as rules over itemsets ordered by time \cite{erminer,rulegrowth,cmrules}, or rules over events in sequences \cite{twincle,spaderulegen,culeassocrulesseq}. The former generally count the number of sequences containing a rule as its support, whereas the latter use sliding or minimal windows to capture multiple rule occurrences within a sequence.

Rules can be further categorized into partially ordered rules \cite{poerm,poermh} where the rule tail follows the rule head but the constituent events of the rule head and the rule tail may appear in any order, and sequential rules where the order both between the rule head and the rule tail as well as within the rule head and the rule tail has to match \cite{spaderulegen}. 
Each of the above approaches consider the quality of individual patterns and hence suffer from the pattern explosion. To address this, Fournier-Viger and Tseng propose \methodname{tns} \cite{tns}, a method that reports the top-k non redundant rules, it uses a strict notion of redundancy and is not able to avoid semantically redundant rules. 

Most closely related to our approach are existing methods that use MDL to select or mine rules \cite{omen2,cuppers2024causal,cossu,comsr}. \methodname{omen} \cite{omen2} is a supervised method for mining `predictive patterns'. It is not applicable in our setting as it requires a target. 
Existing rule set miners for event sequences either filter down an existing set of rules \cite{comsr}, or do not allow for gaps \cite{cossu}. As such, none of the existing methods directly addresses the problem we consider.

\section{Experiments}
\label{sec:exps}

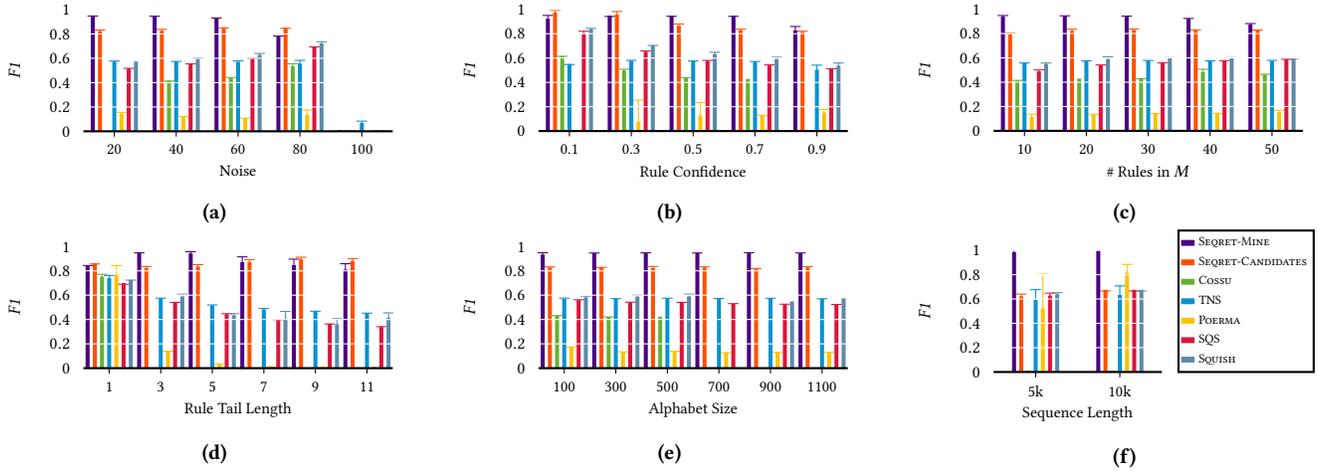
\begin{figure*}
    \begin{center}
        \begin{minipage}{0.32\linewidth}
            \begin{tikzpicture}[outer sep=0pt]
    \begin{axis}[
        pretty ybar,
        xmin = 0.5, 
        xmax = 5.5,
        xtick={1,2,3,4,5},
        xticklabels={20,40,60,80,100},
        xlabel={Noise},
        xlabel style={yshift=-2pt},
        ylabel={$\mathit{F1}$},
        ylabel style={yshift=-34pt},
        legend style = {nodes={scale=1, transform shape}, at = {(1.1,1.06)}, anchor = south},
        legend columns=0,
        height = 3.2cm,
		width = \linewidth,
        bar width = 0.08em,
        ymax=1,
        ymin=0,
    ]
    \pgfplotstableread[col sep=comma]{expres/exp_noise_80_aggregated_results.csv}\tbl

    \foreach \method in {secret,sqsrules_splitgreedy,cossu,tns,poerma,sqs,squish}{  
        \addplot+ [
            unbounded coords=discard,
			error bars/.cd,
			y dir=plus,
			y explicit
            ] table[x=x-,
                    y expr = \strequal{\thisrow{method-}}{\method}?\thisrow{F1-mean}:nan,
                    y error=F1-std
                    ] {\tbl};
    }
        
    \end{axis}
    
\end{tikzpicture}
            \subcaption{}
            \label{fig:exp-noise-level}
        \end{minipage}\hfill
        \begin{minipage}{0.32\linewidth}
            \begin{tikzpicture}[outer sep=0pt]
    \begin{axis}[
        pretty ybar,
        xmin = 0.5, 
        xmax = 5.5,
        xtick={1,2,3,4,5},
        xticklabels={0.1,0.3,0.5,0.7,0.9},
        xlabel={Rule Confidence},
        xlabel style={yshift=-2pt},
        ylabel={$\mathit{F1}$},
        ylabel style={yshift=-34pt},
        legend style = {nodes={scale=1, transform shape}, at = {(1.1,1.06)}, anchor = south},
        legend columns=0,
        height = 3.2cm,
		width = \linewidth,
        bar width = 0.08em,
        ymax=1,
        ymin=0,
    ]
    \pgfplotstableread[col sep=comma]{expres/exp_conf_aggregated_results.csv}\tbl

    \foreach \method in {secret,sqsrules_splitgreedy,cossu,tns,poerma,sqs,squish}{  %
        \addplot+ [
            unbounded coords=discard,
			error bars/.cd,
			y dir=plus,
			y explicit
            ] table[x=x-,
                    y expr = \strequal{\thisrow{method-}}{\method}?\thisrow{F1-mean}:nan,
                    y error=F1-std
                    ] {\tbl};
    }
        
    \end{axis}
    
\end{tikzpicture}
            \subcaption{}
            \label{fig:ep-conf}
        \end{minipage}\hfill
        \begin{minipage}{0.32\linewidth}
            \begin{tikzpicture}[outer sep=0pt]
    \begin{axis}[
        pretty ybar,
        xmin = 0.5, 
        xmax = 5.5,
        xtick={1,2,3,4,5},
        xticklabels={10,20,30,40,50},
        xlabel={\# Rules in $M$},
        xlabel style={yshift=-2pt},
        ylabel={$\mathit{F1}$},
        ylabel style={yshift=-34pt},
        legend style = {nodes={scale=1, transform shape}, at = {(1.1,1.06)}, anchor = south},
        legend columns=0,
        height = 3.2cm,
		width = \linewidth,
        bar width = 0.08em,
        ymax=1,
        ymin=0,
    ]
    \pgfplotstableread[col sep=comma]{expres/exp_num_rules_aggregated_results.csv}\tbl

    \foreach \method in {secret,sqsrules_splitgreedy,cossu,tns,poerma,sqs,squish}{  %
        \addplot+ [
            unbounded coords=discard,
			error bars/.cd,
			y dir=plus,
			y explicit
            ] table[x=x-,
                    y expr = \strequal{\thisrow{method-}}{\method}?\thisrow{F1-mean}:nan,
                    y error=F1-std
                    ] {\tbl};
    }
        
    \end{axis}
    
\end{tikzpicture}
            \subcaption{}
            \label{fig:exp-num-rules}
        \end{minipage}
        \begin{minipage}{0.32\linewidth}
            \begin{tikzpicture}[outer sep=0pt]
    \begin{axis}[
        pretty ybar,
        xmin = 0.5, 
        xmax = 6.5,
        xtick={1,2,3,4,5,6},
        xticklabels={1,3,5,7,9,11},
        xlabel={Rule Tail Length},
        xlabel style={yshift=-2pt},
        ylabel={$\mathit{F1}$},
        ylabel style={yshift=-34pt},
        legend style = {nodes={scale=1, transform shape}, at = {(1.1,1.06)}, anchor = south},
        legend columns=0,
        height = 3.2cm,
		width = \linewidth,
        bar width = 0.08em,
        ymax=1,
        ymin=0,
    ]
    \pgfplotstableread[col sep=comma]{expres/exp_tail_size_aggregated_results.csv}\tbl

    \foreach \method in {secret,sqsrules_splitgreedy,cossu,tns,poerma,sqs,squish}{  
        \addplot+ [
            unbounded coords=discard,
			error bars/.cd,
			y dir=plus,
			y explicit
            ] table[x=x-,
                    y expr = \strequal{\thisrow{method-}}{\method}?\thisrow{F1-mean}:nan,
                    y error=F1-std
                    ] {\tbl};
    }
        
    \end{axis}
    
\end{tikzpicture}
            \subcaption{}
            \label{fig:exp-tail-size}
        \end{minipage}\hfill
        \begin{minipage}{0.32\linewidth}
            \begin{tikzpicture}[outer sep=0pt]
    \begin{axis}[
        pretty ybar,
        xmin = 0.5, 
        xmax = 6.5,
        xtick={1,2,3,4,5,6},
        xticklabels={100,300,500,700,900,1100},
        xlabel={Alphabet Size},
        xlabel style={yshift=-2pt},
        ylabel={$\mathit{F1}$},
        ylabel style={yshift=-34pt},
        legend style = {nodes={scale=1, transform shape}, at = {(1.1,1.06)}, anchor = south},
        legend columns=0,
        height = 3.2cm,
		width = \linewidth,
        bar width = 0.08em,
        ymax=1,
        ymin=0,
    ]
    \pgfplotstableread[col sep=comma]{expres/exp_alph_size_aggregated_results.csv}\tbl

    \foreach \method in {secret,sqsrules_splitgreedy,cossu,tns,poerma,sqs,squish}{  
        \addplot+ [
            unbounded coords=discard,
			error bars/.cd,
			y dir=plus,
			y explicit
            ] table[x=x-,
                    y expr = \strequal{\thisrow{method-}}{\method}?\thisrow{F1-mean}:nan,
                    y error=F1-std
                    ] {\tbl};
    }
        
    \end{axis}
    
\end{tikzpicture}
            \subcaption{}
            \label{fig:exp-alph-size}
        \end{minipage}\hfill
        \begin{minipage}{0.32\linewidth}
            \begin{tikzpicture}[outer sep=0pt]
    \begin{axis}[
        pretty ybar,
        xmin = 0.5, 
        xmax = 2.5,
        xtick={1,2},
        xticklabels={5k,10k},
        xlabel={Sequence Length},
        xlabel style={yshift=-2pt},
        ylabel={$\mathit{F1}$},
        ylabel style={yshift=-34pt},
        legend style = {nodes={scale=1, transform shape}, at = {(1.5,0)}, anchor = south},
        legend columns=1,
        height = 3.2cm,
		width = 3.82cm,
        bar width = 0.08em,
        ymax=1,
        ymin=0,
    ]
    \pgfplotstableread[col sep=comma]{expres/exp_random_1_1_aggregated_results.csv}\tbl

    \addlegendimage{my legend,mambacolor1}
    \addlegendentry{\scriptsize \textsc{Seqret-Mine}}
    \addlegendimage{my legend,internationalorange}
    \addlegendentry{\scriptsize \textsc{Seqret-Candidates}}
    \addlegendimage{my legend,green(ryb)}
    \addlegendentry{\scriptsize \textsc{Cossu}}
    \addlegendimage{my legend,richelectricblue}
    \addlegendentry{\scriptsize \textsc{TNS}}
    \addlegendimage{my legend,goldenpoppy}
    \addlegendentry{\scriptsize \textsc{Poerma}}
    \addlegendimage{my legend,crimson}
    \addlegendentry{\scriptsize \textsc{SQS}}
    \addlegendimage{my legend,airforceblue}
    \addlegendentry{\scriptsize \textsc{Squish}}

    \foreach \method in {secret,sqsrules_splitgreedy,cossu,tns,poerma,sqs,squish}{  %
        \addplot+ [
            unbounded coords=discard,
			error bars/.cd,
			y dir=plus,
			y explicit
            ] table[x=x-,
                    y expr = \strequal{\thisrow{method-}}{\method}?\thisrow{F1-mean}:nan,
                    y error=F1-std
                    ] {\tbl};
    }
        
    \end{axis}
    
\end{tikzpicture}
            \subcaption{}
            \label{fig:exp-random}
        \end{minipage}
    \end{center}
    \caption{[Higher is better] $F1$ scores for synthetic data. We observe that \ourmethod is robust against (a) high noise, (b) rule confidence, (c) number of true rules, (d) rule tail length, and (e) large alphabets. In (f) we evaluate rule recovery where heads and tails are only as frequent as by chance, \ourmethodmine still picks up the ground truth. \out{while \ourmethodcandidates is limited by the input list of patterns}
    }
    \label{fig:}
\end{figure*}

In this section we empirically evaluate \ourmethodcandidates and \ourmethodmine. We implement both in Python and provide the source code, synthetic data generator, and real-world data online.\!\footnote{\oururl} 

We compare \ourmethod to \methodname{poerma}~\cite{poerm} and \methodname{poermh}~\cite{poermh} as representatives of frequent rule mining, to \methodname{tns}~\cite{tns} as a top \(k\) non-redundant rule set miner, to \methodname{cossu}~\cite{cossu} as an MDL-based rule set miner, and to \sqs~\cite{sqs} and \methodname{Squish}~\cite{squish} as MDL-based sequential pattern miners. 

As candidate patterns for \ourmethodcandidates we use the output of \sqs~\cite{sqs} because \methodname{squish} crashes regularly. 
For \methodname{tns} we set \(k\) to the number of rules, and for \methodname{poerma} and \methodname{poermh}, the minimum support and minimum confidence values according to the ground truth when known. In the case of real datasets where the ground truth is unknown, we set $k$ for \methodname{tns} as the number of rules returned by \ourmethodmine. For \methodname{poerma} and \methodname{poermh} we use a minimum support threshold of 10 where feasible, and 20 otherwise. We allow all methods a maximum runtime of 24 hours. With the exception of \cossu, which we allow a maximum runtime of 48 hours, as it generally took longer to complete. We provide additional details in Appx.~\ref{apx:real-world-table}.

\subsection{Synthetic Data}
We first consider data with known ground truth. To this end, we generate synthetic data from a randomly generated rule set $R$. For a given alphabet size, number of rules, sizes of the heads and tails, and confidence, we generate rule heads and the rule tails by selecting events from alphabet $\Omega$ uniformly at random with replacement.

\paragraph{Generating Data} We generate event sequences $S$ as follows. We first generate background noise by sampling uniformly at random from the alphabet. Next, we plant patterns, i.e. the empty-head rules from $R$. We sample uniformly from all empty-head rules in the model and write the tails to $S$ at random positions while making sure we do not overwrite existing rules. Finally, we go over the generated sequence and wherever a non-empty-head rule is triggered, we sample according to the desired confidence of the rule whether the trigger is a hit or a miss. If it is a hit, we sample the delay and then insert the corresponding rule tail. For all tails we so plant, we sample gap events too. The probability of delay and gaps are set as input parameters. We provide additional details of synthetic data generation in Appx.~\ref{apx:sec:synth}.

Unless stated otherwise, we generate sequences of length $10\,000$ over alphabets of size $500$, rule sets of size $20$ with rule confidence $0.75$. We generate 20 datasets per configuration in each experiment. 

\paragraph{Evaluation Metric}
As evaluation metric, we consider the $F1$ score. To reward partial discovery, we compute recall and precision based on the similarity between the discovered rules and the ground truth rules. We evaluate the similarity between any two rule heads or rule tails using the Levenshtein edit distance without substitution, i.e the longest common subsequence distance \cite{lcs}. To keep the similarities comparable between rules we normalize by the combined lengths. Formally, we have 
$\rulesimilarity{X}{U} =   1 - \mathit{lcsd}(X,U)/(|X|+|U|)$, where $\mathit{lcsd}(X,U) = |X|+|U|-2|lcs|$. 
To reward recall of rule heads, rule tails, and complete rules, we evaluate each similarity between all parts separately. The similarity between two rules is then a weighted average, $\rulesimilarity{X \rightarrow Y}{U \rightarrow V} = \rulesimilarity{XY}{UV}/2 + \rulesimilarity{X}{U}/4 + \rulesimilarity{Y}{V}/4$.
We then compute recall and precision following C\"{u}ppers et al.~\cite{omen1}. To compute recall, we sum the highest similarity per rule in the true model for the mined rules, and divide the sum by the number of rules in the true model. To compute precision, to avoid rewarding redundancy, we limit the sum to $T$ rules with highest similarity scores, where $T$ is the number of rules in the rule model. We provide additional details in Appx.~\ref{apx:sec:eval}.

\paragraph{Sanity Check}  
We first evaluate if our score indeed prefers rules over other patterns. To this end, we generate synthetic data using a ground truth rule set consisting of 6 pairs of the form \(\set{\epsilon \rightarrow X, X \rightarrow Y}\). We then compare the encoded sizes of the ground truth model against alternative models of the form \(\set{\epsilon \rightarrow X, \epsilon \rightarrow Y}, \set{\epsilon \rightarrow XY} \) resp. \(\set{\epsilon \rightarrow X, \epsilon \rightarrow XY}\). We find that our score always prefers the ground truth. Next, we evaluate on data without structure. We find that for 60 trials on sequences varying in size from $5\,000$ to $15\,000$, \ourmethodmine, in $51$ instances correctly reports no rules. In 9 instances, it returns a single, rule with a true confidence between 10\% and 40\%.

\paragraph{Destructive Noise}
Next, we evaluate robustness against destructive noise. To this end, we generate data as above and then add noise by flipping individual events $e \in S$ with probability ranging from 20\% to 100\%. We show the results in Figure \ref{fig:exp-noise-level}. We observe that \ourmethod is robust against noise and still recovers the ground truth well even at 80\% noise. At 100\% noise, there is no structure in the data and all methods except \methodname{tns} correctly discovers no rules. Throughout all synthetic experiments \poerma performs better than \poermh, we hence omit \poermh from the synthetic experiments results. Further, as \methodname{squish} regularly crashes, we report averages over finished runs only. 

\begin{table*}[t]
    \centering
    \begin{tabular*}{0.95\linewidth}{@{\extracolsep{\fill}} l rrr@{\hspace{1em}} rrr@{\hspace{1.5em}} rrr rrrrrrr}
      \toprule
      \multicolumn{4}{l}{} & \multicolumn{3}{l}{\methodname{\ourmethod-Cands}} & \multicolumn{3}{l}{\ourmethod-\methodname{Mine}} & \multicolumn{2}{l}{\methodname{sqs}} & \poerma & \poermh & \tns & \multicolumn{2}{l}{\cossu} \\
      \cmidrule(r){5-7}
      \cmidrule(){8-10}
      \cmidrule(){11-12}
      \cmidrule(){13-13}
      \cmidrule(){14-14}
      \cmidrule(){15-15}
      \cmidrule(){16-17}
      \textbf{Dataset} & $||D||$ & $|D|$ & $|\Omega|$ &  $|P|$ & $|R|$ & $\%L$ & $|P|$ & $|R|$ & $\%L$ &  $|P|$ & $\%L$ & $|R|$ & $|R|$ & $|R|$ & $|R|$ & $\%L$ \\
      \midrule
 {Ordonez}     & 739 & 2    & 10   & 2 & 0  & 11.45  & 1  & 5  & 16.85  & 2  & 11.45  & 95923 & 113337 & 6 & 2 & -2.42 \\
      JMLR               & 14501 & 155  & 1920 & 62 & 9  & 1.90 & 2  & 203  & 3.33 & 116 & 1.61 & 127 & 1282442 & 205 & -- & --  \\
 {Rollingmill} & 18416 & 350  & 446  & 158 & 50 & 52.22 & 9  & 313 & 56.33 & 247 & 50.46 & -- & -- & 247 & 46  & 20.56 \\
{Lichess}     & 20012 & 350  & 2273 & 81 & 18 & 2.42 & 11 & 326 & 4.57 & 113 & 2.13 & 4326  & 2068 & 348 & -- & -- \\      
{Ecommerce}   & 30875 & 4001  & 127  & 77 & 10  & 13.72 & 89  & 130 & 27.87 & 95  & 13.59 & 43513 & 231824 & 219 & 6 & -0.09 \\
 {Lifelog}     & 40520 & 1  & 78   & 36 & 6 & 8.85  & 16 & 79 & 9.97  & 59 & 6.51  & 2001521 & 1932296 & 95 & 5 & -1.66  \\
 {POS}         & 45531 & 1761  & 36   & 65 & 6 & 18.36 & 38 & 33  & 18.12 & 160 & 12.64 & -- & -- & 71 & 5  & -0.52 \\
 {Presidential} & 62010 & 30  & 3973 & 30 & 4  & 0.46 & 2  & 129 & 0.96 & 58  & 0.38 & 57 & 90552 & 131 & -- & --   \\
      \bottomrule
    \end{tabular*}

    \caption{
        Results on real-world data. We report the number of discovered patterns (non-empty-head rules) $P$ and rules $R$. For \ourmethod, \methodname{sqs} and \methodname{cossu}, we report the percentage of bits saved against the \ourmethod null model as $\%L$. Failed runs, e.g. because of excessive runtime (\methodname{cossu}) or out-of-memory errors (\methodname{poerma} and \methodname{poermh}), are indicated by `--'. 
    }
    \label{tbl:real-world-overview}
\end{table*}

\paragraph{Rule Confidence}
Next, we evaluate recovery under different rule confidence levels. We vary the rule confidence from 0.1 to 0.9. We show the results in Figure \ref{fig:ep-conf}. We observe that \ourmethod is robust against low confidence rules and outperforms the state-of-the-art methods by a clear margin.

\paragraph{Varying Size}
Next, we evaluate how \ourmethod performs for ground truth models, alphabets, resp. rule tails of different sizes. First, we consider data with 10 to 50 ground truth rules. We show the results in Fig.~\ref{fig:exp-num-rules}. \ourmethod works consistently well across all model sizes.
Next, we vary the alphabet from $100$ to $1000$ unique events. We give the results in Fig.~\ref{fig:exp-alph-size} and observe that \ourmethod recovers the ground truth consistently well. Finally, we vary the length of the rule tails from 1 to 11. We show the results in Figure \ref{fig:exp-tail-size}. We observe that, except for the case where the rule tail size is 1, \ourmethod recovers the rules well and outperforms the competition by a large margin.

\paragraph{Random Rule Triggers}
Finally, we evaluate if \ourmethod can recover conditional dependencies even when the rule heads and rule tails are infrequent in the data. For this, we consider the case where the rule heads occur only by chance.
To this end, we generate synthetic data where no rule heads are planted. We insert the rule tails wherever the corresponding rules have triggered. To ensure that the rule heads do occur in the data, we limit its size to 1. We also limit the size of the tail to 1 to ensure they do not stand out as patterns by themselves. We show the results in Figure \ref{fig:exp-random}. \ourmethodmine is able to consistently recover the rules. Here, \sqssqrt performs significantly worse because \sqs is not able to find good patterns in this challenging setting.

\subsection{Experiments on Real Datasets}

In this section, we examine if \ourmethod mines insightful rules from real world data. We first discuss the datasets and then the results.

\paragraph{Datasets} 
We use eight datasets drawn from five different domains. We consider two text datasets, \emph{JMLR}, which contains abstracts from the JMLR journal, and \textit{Presidential}, which contains addresses delivered by American presidents~\cite{sqs}. \textit{POS} contains sequences of parts-of-speech tags obtained by using the Stanford NLP tagger on the book ``History of Julius Caesar'' by Jacob Abbott \cite{pos}. \textit{Ordonez} \cite{ordonez2013activity} and \textit{Lifelog}\footnote{https://quantifiedawesome.com/} contains the daily activities logged by a person over several days. \textit{Rolling Mill} contains the process logs from a steel manufacturing plant \cite{proseqo}. 
\textit{Ecommerce} contains the purchase history from an online store for several users over 7 months.\!\footnote{\href{https://www.kaggle.com/datasets/mkechinov/ecommerce-behavior-data-from-multi-category-store}{https://www.kaggle.com/datasets/mkechinov/ecommerce-behavior-data-from-multi-category-store}} 
Finally, \textit{Lichess} contains sequences of moves from chess games played online.\!\footnote{\href{https://www.kaggle.com/datasets/datasnaek/chess}{https://www.kaggle.com/datasets/datasnaek/chess}} In Table \ref{tbl:real-world-overview} we provide statistics on all datasets, as well as on the results of the different methods.

\paragraph{General Observations}
Overall, we observe that frequency-based methods like \poerma and \poermh discover a high number of rules, making interpretation difficult up to impossible. \tns produces largely redundant rules. \cossu is limited by its restrictive rule language and discovers only very few rules.
\sqs is a sequential pattern miner that identifies meaningful patterns that permit examination by hand, but does not capture conditional dependencies that \ourmethod does successfully model. This difference is evident when comparing the compression achieved by different methods: \sqs compresses the data less effectively than \ourmethod, likely because \ourmethod is more expressive. \sqssqrt improves upon \sqs but still compresses worse than \ourmethodmine.

\paragraph{Case Studies}
Next, we present illustrative examples to highlight how the results from \ourmethod differ from those of state-of-the-art methods.
To better understand the results, we examine a phrase from the \jmlr dataset, `\emph{support vector machine}', that all methods identify in some form.  
\ourmethodmine discovers the rule \emph{< $ \epsilon \rightarrow$ support, vector>}  as well as the rule \emph{<support, vector} $\rightarrow$ \emph{machine>} which expresses that \emph{<support}, \emph{vector>} is a pattern, and that whenever it occurs it increases the probability of but is \emph{not necessarily} followed by $\emph{machine}$. In contrast, \sqssqrt and \sqs both treat the entire phrase as a single pattern, \emph{<support, vector, machine>}, failing to capture the independent existence of \emph{<support, vector>} and the conditional dependency involved. 
On the other end of the spectrum, \poerma and \tns discover 12 resp. 14 rules involving either $\emph{support}$ or $\emph{vector}$, many of which are semantically redundant.

In the \textit{POS} dataset, \ourmethod discovers common sentence structures, such as the pattern \emph{<to}, \emph{verb-base-form>}, and the pattern \emph{<determiner}, \emph{cardinal number>} capturing phrases such as ``the first''.
\ourmethod also captures rules, e.g. \emph{<to}, \emph{verb-base-form} $\rightarrow$ \emph{personal-pronoun>} and \emph{<to}, \emph{verb-base-form} $\rightarrow$ \emph{possessive-pronoun>}. These rules correctly identify how either personal pronouns or possessive pronouns tend to follow phrases like ``to tell'' or ``to give''. \methodname{cossu}, the method closest to our approach, fails to find any of the rules discussed.  Meanwhile,  \methodname{sqs} finds these structures as several independent patterns disregarding the conditional dependency. The frequency-based methods again return overly many and highly redundant rules. 

On the \textit{Rolling Mill} data, \ourmethod discovers rules that clearly represent different parts of the production process. For example, \emph{<stab}, \emph{gies}, \emph{sort} $\rightarrow$ \emph{brwa}, \emph{lmbr}, \emph{tmbr>} captures the transition from steel mill (where hot iron is casted and sorted to slabs) to the rolling mill (where slabs are rolled to plates).
This demonstrates the power of rules and patterns in the same set, as there are instances where \emph{<stab}, \emph{gies}, \emph{sort>} is not followed by \emph{<brwa}, \emph{lmbr}, \emph{tmbr>} but such instances are rare. \methodname{sqs} again finds several patterns involving parts of \emph{<stab}, \emph{gies}, \emph{sort}, \emph{brwa}, \emph{lmbr}, \emph{tmbr>} but does not explicitly model the conditional dependency. \cossu fares better than in other datasets but nevertheless misses many important dependencies resulting in poor compression.

\begin{figure}
    \begin{center}
        \begin{minipage}{0.49\linewidth}
            \centering
            \includegraphics[width=0.7\linewidth]{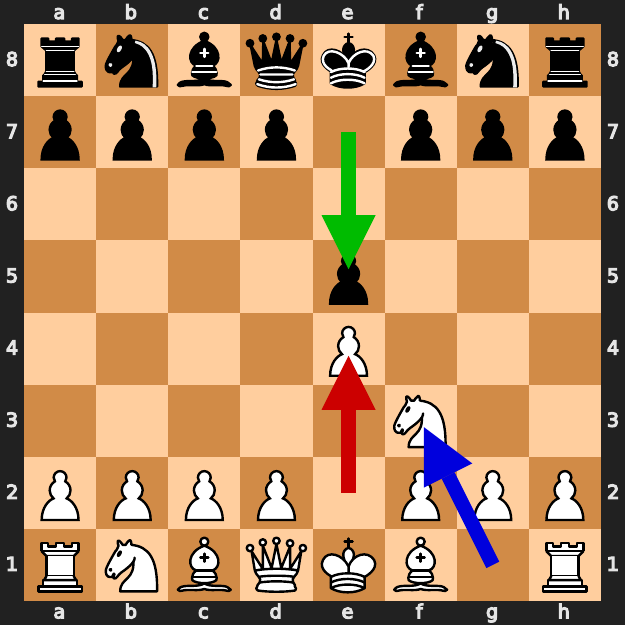}
            \subcaption{}
            \label{fig:chess-kings-pawn}
        \end{minipage}
        \hfill
        \begin{minipage}{0.49\linewidth}
            \centering
            \includegraphics[width=0.7\linewidth]{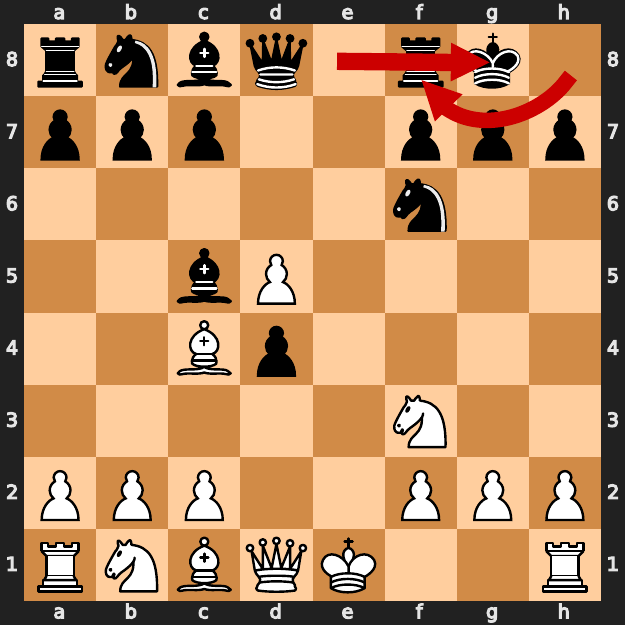}
            \subcaption{}
            \label{fig:chess-castle}
        \end{minipage}
        \hfill
    \end{center}
    \caption{Rules discoverd on the \textit{Lichess} dataset:  In (a) we show the rule \emph{<white:e4 (red arrow), black:e5 (green arrow)} $\rightarrow$  \emph{white:Nf3 (blue arrow)>}. In (b) we show black castling \emph{<black:O-O>}. Castling is a special chess move where the king moves two squares toward a rook, and the rook jumps over the king to the adjacent square.}
    \label{fig:chess}
\end{figure}

For the \textit{Lichess} dataset, \ourmethod finds the well known ``King's Pawn Game'' opening move, as the pattern \emph{<white:e4}, \emph{black:e5>}. In addition, it discovers 12 rules with rule head \emph{<white:e4}, \emph{black:e5>}, capturing the different variations that often follow. For example the rule \emph{<white:e4}, \emph{black:e5} $\rightarrow$ \emph{white:Nf3>}, we show this rule in Figure \ref{fig:chess-kings-pawn}, the red arrows corresponds to move \emph{<white:e4>}, green to \emph{<black:e5>}, and blue to \emph{<white:Nf3>}. 
\sqs, on the other hand, needs several partly redundant patterns, i.e. repeating moves \emph{<white:e4, black:e5>} and \emph{<white:Nf3>}, to explain the same dependencies. 

Diving deeper into results on \textit{Lichess}, \ourmethod discovers rules involving ``King's side castling" (denoted by O-O and shown in Figure \ref{fig:chess-castle}) which despite being insightful conditional dependencies are missed by \sqs.
Examples are \emph{<black:O-O} $\rightarrow$ \emph{black:Re8>} and \emph{<black:O-O} $\rightarrow$ \emph{white:Qe2>}. The former captures black moving its rook to e8, a position originally occupied by the King and made available only after castling. The latter captures white moving its queen to e2 following black castling, as a deterrence to black rook (as e8 is in the line of attack of the queen). Another example is the pattern \emph{<black:Nf6}, \emph{black:O-O>} which makes sense as moving the knight away is a prerequisite for castling. For frequency-based methods, we find anywhere between 128 and 1370 rules involving castling, most of which are redundant.

\section{Conclusion}\label{sec:conclusion}

We considered the problem of mining a succinct set of rules from event sequences. We formalized the problem in terms of the MDL principle and presented the \ourmethodcandidates and \ourmethodmine algorithms.
We evaluated both on synthetic and real-world data. On synthetic data we saw that \ourmethod recovers the ground truth well and is robust against noise, low rule confidence, different alphabet sizes, and rule set sizes. On real-world data \ourmethod found meaningful rules and provides insights that existing methods cannot provide. 

As future work, we consider it highly interesting to study the causal aspects of sequential rules. Our approach lends itself to a causal framework by mapping the rule heads and tails to temporal variables and re-modeling the rules as structural equations involving these variables. Moreover, the MDL principle as used in this paper nicely maps to the Algorithmic Markov Condition \cite{amc} criterion to choose a plausible causal model.

	\bibliographystyle{abbrv}
	\bibliography{bib/abbrev,bib/bib-jilles,bib/bib-paper,bib/references,bib/transactional,bib/causal,bib/sequential}
	
	\ifapx
	\appendix
	\onecolumn
	\section*{Appendix}
\label{sec:apx}

\section{Supplementary Material: Experiments}
\label{apx:sec:synthevalres}

\subsection{Setup}
\label{apx:real-world-table}
We ran all experiments on an Intel Xeon Gold 6244 @ 3.6 GHz, with 256GB of RAM. 
For the methods \poerma and \poermh, the JVM max heap size was increased upto 64 GB. Both these methods discover partially ordered rules from long event sequences, albeit with different definitions of rule support. To ensure fair comparison, we constraint our synthetic data generation to rules where constituent events appear in lexicographical order, and re-arrange the partially-ordered rules found to this order.
We set a time limit of 24 hours for all methods except \methodname{cossu}. As \cossu took a very long time to complete across all experiments, we increased the time limit for \cossu alone to 48 hours. Across all synthetic experiments except where the rule tail size was varied, \ourmethodmine completed within a few seconds to 1 hour max. In experiment varying rule tail size, \ourmethodmine took upto 3 hours in a few instances indicating that data with highly interleaved rules take up more time in rule search. We report the runtimes on real datasets in Table \ref{tab:realruntime}.

\subsection{Synthetic Data Generation}
\label{apx:sec:synth}
Given an alphabet as input, we first generate a random rule set. We take in as parameters the rule set size, the rule-head size, the rule-tail size and the rule confidences. We also parameterize whether or not the rule heads occur as independent patterns, i.e for a rule \(X \rightarrow Y\), whether \(\epsilon \rightarrow X\) exists or not. If \(\epsilon \rightarrow X\) doesn't exist, then \(X\) is only as frequent as expected by chance. Given these parameters, we randomly select events from the alphabet to form the rule heads and the rule tails, and add them to the rule set.

Next, using the rule set as ground truth, we generate the sequence database. We first generate an initial sequence using all the empty-head rules, and then insert the rule tails wherever the non-empty-head rules have triggered. We take in as parameters an initial sequence size and noise percentage. By noise, we mean the events in the sequence that can be covered only using one of the singleton rules. Therefore, given a noise percentage, we uniformly sample from the singleton rules, i.e the alphabet, to generate the stipulated percentage of the initial sequence size. Following this, we uniformly sample from the empty-head non-singleton rules and fill them into random positions to generate the remaining sequence. Finally, we go over the generated sequence, identify rules that have been triggered and insert the corresponding rule tails as per the specified rule confidences.

As for the delays and gaps, we take in as parameters a delay probability, i.e probability with which the data generation algorithm skips positions following a rule trigger, and a gap probability, i.e probability with which the data generation algorithm skips positions within rule tails. Note that the insertion of rule tails will alter the sequence size and the noise percentage. We keep the gap and delay probabilities low at 0.1 and 0.2 respectively. To run \ourmethod, we set the \maxdelay and the \maxgap both to 2.

\subsection{Evaluation Metrics}
\label{apx:sec:eval}
To evaluate a method, we compare the rule set retrieved by it the against the ground truth in terms of recall and precision. To measure the similarity between rules quantitatively, we define a metric based on the LCS distance measure \cite{lcs}. As the LCS distance is upper bounded by the sum of lengths of the patterns and lower bounded by zero, we can compute the similarity between any two patterns \(A\) and \(B\) as \[\patternsimilarity{A}{B} = 1 - \frac{\LCSdistance{A}{B}}{|A| + |B|}\;,\] where \(\LCSdistance{A}{B}\) refers to the LCS distance. Further, we define the similarity measure between two rules \(A \rightarrow B\) and \(C \rightarrow D\) where either \(A\) or \(C\) is non-empty as

\begin{equation}
    \begin{aligned}
        \rulesimilarity{A \rightarrow B}{C \rightarrow D} \;=\; &0.5 * \patternsimilarity{AB}{CD} \\ +\; &0.25 * \patternsimilarity{A}{C} \\ +\; &0.25 * \patternsimilarity{B}{D}
    \end{aligned}
\end{equation}

Using this similarity measure, we calculate recall and precision. To compute recall, we sum up the similarity measure of each true rule with respect to the best matching mined rule, and normalize the sum. Formally, given a true model \(T\) and a mined model \(M\), we define 
\begin{equation}
    \recall{T}{M} = \frac{\sum_{t \in T} \max_{m \in M} \rulesimilarity{t}{m}}{|T|}\;.   
\end{equation}

Similarly, to compute precision, we use the similarity measures of the mined rules with respect to their best matching true rules. To penalize redundancy in the mined model, we should ideally choose a non-redundant subset of the mined model that maximizes the total similarity measure. However, this is not trivial to solve. Therefore, we resort to a heuristic measure as proposed in \methodname{omen} \cite{omen1}, and choose the top \(|T|\) maximum similarity measures from the mined model. Given true model \(T\) and mined model \(M\),
\begin{equation}
    \precision{T}{M} = \frac{\sum_{m' \in M'}m'}{|M|}\;,
\end{equation}
where 
\(
M' \subseteq \{\max_{t \in T} \rulesimilarity{t}{m} \mid \forall m \in M\}
\)
such that \(M'\) contains the max \(|T|\) elements. Finally, we use precision and recall to compute the F1 score.

\begin{table}[h]
    \centering
    \begin{tabular}{|c|c|}
        \hline
        Dataset & Runtime \\ 
        \hline
        \jmlr & 4 h \\
        \textit{Presidential} & 8 h \\
        \textit{POS} & 3 h \\
        \textit{Lifelog} & 2 h \\
        \textit{Ordonez} & 3 s \\
        \textit{Ecommerce} & 11 h \\
        \textit{Rollingmill} & 22 h \\
        \textit{Lichess} & 17 h \\
        \hline
    \end{tabular}
    \caption{Runtime of \ourmethodmine for different datasets.}
    \label{tab:realruntime}
\end{table}

\section{Supplementary Material: Algorithms}
\label{apx:sec:algos}
\subsection{Rule Windows}
\label{apx:sec:rule-window}

A rule window \(S[i,j;k,l]\) captures the positions at which a rule occurs in a sequence \(S\). Here, \(S[i,j]\) is the window within which the rule head occurs and \(S[k,l]\) is the window within which the rule tail occurs such that \(j \leq k\). To avoid double counting and minimize gaps, we use minimal windows to identify the rule head patterns that trigger the rules. But what about the rule windows? For each rule head, we prefer the nearest minimal window of the rule tail pattern to complete the rule window. We prefer minimal windows as they minimize the gaps and treat the rule tail as a cohesive unit. If multiple minimal windows of the rule tail exist following the trigger, then we pick the nearest one so as to minimize the delay. Further, we restrict the search to windows that follow a user-set \maxdelay ratio such that ${k-j-1}/{\vert \rtaill{r}\vert} \leq \maxdelay$, and a \maxgap ratio such that ${l-k+1}/{\vert \rtaill{r}\vert} \leq \maxgap$ and ${j-i-1}/{\vert \rheadd{r}\vert} \leq \maxgap$.

Algorithm \ref{alg:findwin} outlines the pseudo-code to find the best rule windows for a given rule. 

The \bestrwins method, however, assumes that none of the events forming the preferred windows for different rules have already been covered. As we start covering the sequence with these windows, however, it may happen that, at some point, events that are common to multiple rule tails have already been covered. In such cases, we look for the next best rule window. The next best rule window is the nearest minimal window of the rule tail following the trigger such that the events forming the rule tail are not already covered. Algorithm \ref{alg:findnextwin} outlines the pseudo-code for the \nextrwin method.  

As an example, consider the sequence \(\langle a,a,b,b,c,c,d,d\rangle\) in Figure \ref{fig:examplerwins} and rule \(ab \rightarrow cd\). The minimal window \(S[1,2]\) captures the rule head that triggers the rule. Assuming none of the events in the sequence have already been covered, we pick \(S[5,6]\) as the rule tail window. The alternate windows for the rule tail are considered if and only if positions 5 or 6 have already been covered by other rules.

\begin{figure}
\centering
\includegraphics[width=0.5\textwidth]{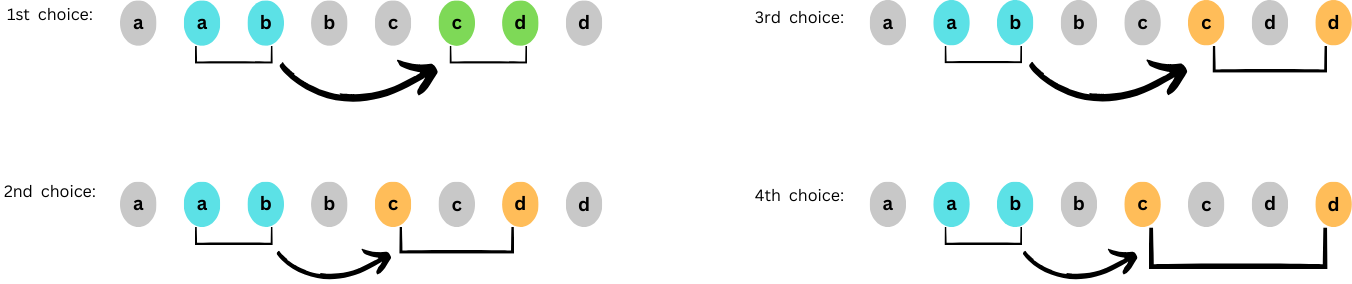}
\caption{An illustration of potential rule windows for the rule \(ab \rightarrow cd\). We pick the nearest minimal window of the rule tail as our preferred window.}
\label{fig:examplerwins}
\end{figure} 

\begin{algorithm}[h]
\caption{\bestrwins}
\label{alg:findwin}
\SetAlgoNoEnd
\SetAlgoLined
\SetCommentSty{mycommfont}
\SetKwRepeat{Do}{do}{while}
\KwIn{\(\varname{rule},D\)}
\KwOut{\(\varname{windows}\)}
$\varname{windows} \gets \set{}$\;
\(
\varname{triggers} \gets \set{S[i,j] \mid S \in D,\; (j-i+1) - \vert\rheadd{\varname{rule}}\vert \leq \maxgap * \vert\rheadd{\varname{rule}}\vert}
\;\), \\ \hspace{0.4em} where \(S[i,j]\) is a minimal window of \rheadd{\varname{rule}}\;
\For{$S[i,j] \in \varname{triggers}$}{
     $k,l \gets$ indices of first minimal window $S[k,l]$ of \rtaill{\varname{rule}} such that $(k-j-1) \leq \maxdelay * \vert\rtaill{\varname{rule}}\vert$ and $(l-k+1) - \vert\rtaill{\varname{rule}}\vert \leq \maxgap * \vert\rtaill{\varname{rule}}\vert$\;
     \If{$S[k,l]$ exists}{
        $\varname{windows} \gets \varname{windows} \cup \set{(\varname{rule},S[i,j;k,l])}$\;
     }
}
\Return \varname{windows}
\end{algorithm}

\begin{algorithm}[h]
\caption{\nextrwin}
\label{alg:findnextwin}
\SetAlgoNoEnd
\SetAlgoLined
\SetCommentSty{mycommfont}
\SetKwRepeat{Do}{do}{while}
\KwIn{\(\varname{win}, \varname{cover}, D\)}
\KwOut{\(\varname{win'}\)}
\tcc{\varname{win} contains a pointer to the rule (\varname{win.rule}) and the window in the format \(S[i,j;k,l]\)}
\(k',l' \gets\) indices of first minimal window $S[k',l']$ of \rtaill{\varname{win.rule}} such that $\forall e \in \set{e \mid S[k',l'] \;\text{covers}\; e}$, $\nexists w \in \varname{cover}$ where $w$ covers $e$ and
$(k'-j-1) \leq \maxdelay * \vert\rtaill{\varname{win.rule}}\vert$ and $(l'-k'+1) - \vert\rtaill{\varname{win.rule}}\vert \leq \maxgap * \vert\rtaill{\varname{win.rule}}\vert$\;
$\varname{win'} \gets (\varname{win.rule},S[i,j;k',l'])$\;
\Return \varname{win'}
\end{algorithm}

\subsection{Prune Algorithm}
The Algorithm \ref{alg:prune} shows the pseudo-code for the \methodname{Prune} procedure.
\begin{algorithm}[h]
\caption{\prune}
\label{alg:prune}
\SetAlgoNoEnd
\SetAlgoLined
\SetCommentSty{mycommfont}
\SetKwRepeat{Do}{do}{while}
\KwIn{\(D,R\)}
\KwOut{pruned \(R\)}
\For{$r \in R$ ordered by \pruneorder}{
    \If{\(r\) is not a singleton rule}{
        \If{$L(D,R\setminus \set{r}) < L(D,R)$}{
            $R \gets R\setminus \set{r}$\;
        }
    }
}
\Return \(R\)
\end{algorithm}

\subsection{CandRules Algorithm}
\label{apx:sec:candidates}
The Algorithm \ref{alg:neigh} shows the pseudo-code for the \methodname{CandRules} procedure.
\begin{algorithm}[h]
\caption{\signineigh}
\label{alg:neigh}
\SetAlgoNoEnd
\SetAlgoLined
\SetCommentSty{mycommfont}
\SetKwRepeat{Do}{do}{while}
\KwIn{\(D, \Omega, \varname{rule}\)}
\KwOut{\(\varname{candidates}\)}
$\varname{candidates} \gets \set{}$\;
$\varname{windows} \gets \bestrwins(\varname{rule},D)$\;
\For{$\varname{position} \in \set{h_0,h_1,...,h_{\vert\rheadd{\varname{rule}}\vert}} \cup \set{t_0,t_1,...,t_{\vert\rtaill{\varname{rule}}\vert}}$}{
    \tcc{$h$ represents rule head and $t$ represents rule tail}
    \For{$e \in \Omega$}{
        $\varname{count} \gets \vert\set{w \in \varname{windows}\mid w \;\text{contains}\; e \;\text{in the gap at}\; \varname{position}}\vert$\;
        $p_{e^{c}} \gets 1 - \frac{\supp{\epsilon \rightarrow e}}{\vert D \vert}$\tcc*[l]{$e^{c}$ refers to the complement of $e$}
        $\varname{expected} \gets  \vert \varname{windows}\vert - \sum_{w \in \varname{windows}}{(p_{e^{c}})^{\vert g_w\vert}}$\;
        \tcc{$g_w$ refers to the gap in $w$ at \varname{position}}
        \If{\varname{count} significantly greater than \varname{expected}}{
            \tcc{see section \ref{solgreedyminer} for details of significance test}
            $\varname{candidates} \gets \varname{candidates} \cup \set{\methodname{Insert}(\varname{rule}, e, \varname{position})}$ 
            \tcc{\methodname{Insert} inserts $e$ to \varname{rule} at \varname{position}}
            }
    }
}
\Return \varname{candidates} 
\end{algorithm}

\subsection{Split Algorithm}
The Algorithm \ref{alg:split} shows the pseudo-code for the \methodname{Split} procedure.
\begin{algorithm}[tb!]
    \caption{\textsc{Split}}
    \label{alg:split}
    \KwIn{Pattern $p$}
    \KwOut{Set of rules $R$}
        $R \leftarrow \emptyset$\\
        $i \leftarrow 0$ \\
        \While{$i < |p|$}{
            $R \leftarrow R \cup {(p[0,i],p[i+1,|p|])}$\\
            $i \leftarrow i+1$\\
        }
    \Return{$R$}
\end{algorithm}

\section{Time Complexity Analysis}
\subsection{Time Complexity of the Rule-Set Mining Problem}
\label{ogcomplexity}

To evaluate the time complexity of the problem, let us split the problem into two parts - one, to find the optimal cover given a rule set, and two, to find the optimal rule set. For simplicity, let us assume a single long sequence \(S\) in the database, drawn from the alphabet \(\Omega\). 

Given a rule set \(R\), we know that a cover is a many-to-one mapping between the events in \(S\) to rules in \(R\). In other words, it is a permutation with replacement of the rules in \(R\) over \(\vert S \vert\) events. Therefore, we can compute the number of possible covers as 
\( \vert R \vert^{\vert S \vert}\). The worst-case time complexity of the first part of our problem is 
\[ \mathcal{O}(\vert R \vert^{\vert S \vert})\;.\]
Now let us compute the time complexity of the second part of our problem. The longest rule that can occur in \(S\) would be of length \(\vert S \vert\). The total number of rules possible would be the sum of the number of rules possible per size, with size ranging from 1 to \(\vert S\vert\). Considering that rules can be built from sequential patterns, let us first compute the number of sequential patterns possible for size \(k\). A sequential pattern is a permutation of the alphabet with replacement. Therefore, for size \(k\) we get \(\vert \Omega \vert^k\) possible sequential patterns. Now, including the possibility of an empty-head, we can choose \(k\) positions to split the pattern into a rule head and a rule tail. Thus, for size \(k\), we can compute the number of rules possible as \(k * \vert \Omega \vert^k\). The total number of rules possible is then given by
\[ \sum_{k = 1}^{\vert S\vert} k * \vert \Omega \vert^k\;.\]
A rule set being a subset of all possible rules, we can compute the number of possible rule sets as the size of the power-set. Since we retain the singleton rules in every possible rule set, we can compute the number of valid rule sets as \(2^{\sum_{k = 1}^{\vert S\vert} k * \vert \Omega \vert^k - \vert \Omega \vert}\). The worst-case time complexity of the second part of our problem is then given by 
\[\mathcal{O}(2^{\sum_{k = 1}^{\vert S\vert} k * \vert \Omega \vert^k - \vert \Omega \vert}) \;\approx\; \mathcal{O}(2^{\vert \Omega \vert^{\vert S \vert}})\;.\]

\subsection{Time Complexity of \ourmethodmine}
\label{solcomplexity}

Let us now analyze the time complexity of our solution. We will analyze each part of the problem separately. We consider the worst-case where all the events in the database occur as a single long sequence \(S\). The set of distinct events form the alphabet \(\Omega\).

\paragraph{Time Complexity of \cover}
Given a rule set \(R\), we first compute the complexity of finding the rule windows. To do so, the method looks for all rule triggers and for each rule trigger, finds the nearest minimal window of the rule tail. Iterating over \(S\) and looking for triggers of each rule \(r \in R\) results in a worst-case time complexity of \(\mathcal{O}(\vert S \vert * \sum_{r \in R} \vert \rheadd{r} \vert * \maxgap)\). Ignoring the \maxgap parameter that stays constant irrespective of the problem size and upper bounding the size of any rule head by \(\max_{r \in R} \vert \rheadd{r}\vert\), denoted by \varname{max\_head\_size}, we can rewrite the same as \[\mathcal{O}(\vert S \vert * \vert R\vert * \varname{max\_head\_size})\;.\] 
To complete the rule window for each trigger, \bestrwins next looks for the nearest minimal window of the rule tail until the maximum allowed delay. Given a trigger, looking for a rule tail, for any rule \(r\), requires computational time in the order of 
\(
\mathcal{O}(\vert \rtaill{r} \vert * \maxdelay * \vert \rtaill{r} \vert * \maxgap)\). Once again ignoring the constant parameters and using \varname{max\_tail\_size} to upper bound the size of a rule tail, we can rewrite this as
\(\mathcal{O}(\varname{max\_tail\_size}^2)\).
Thus, we can compute the worst-case time complexity of \bestrwins as 
\begin{equation}
\label{appeqnfrw}
    \mathcal{O}\left(\vert S \vert * \vert R \vert * (\varname{max\_head\_size} + \varname{max\_tail\_size}^2)\right)\;.
\end{equation}
As triggers are bounded by minimal windows, and only one minimal window can exist per starting or ending position, the number of triggers per rule is upper bounded by \(\vert S \vert\). Since \bestrwins finds only the one nearest minimal window of the rule tail for each trigger, we can upper bound the total number of rule windows returned to \(\vert R\vert * \vert S\vert\). The next step in \cover is to sort the rule windows in \windoworder. This incurs a time complexity of 
\begin{equation}
\label{appeqnsrt}
   \mathcal{O}(\vert R\vert * \vert S\vert * \log_2(\vert R\vert * \vert S\vert))\;. 
\end{equation}
The final step in \cover is to consider each rule window in the sorted order and cover the sequence. However, there could arise cases where the considered rule windows are in conflict with previous rule windows which already covered the same events. This in turn leads to the execution of \nextrwin. Each time \nextrwin is called for a rule trigger, it looks for the next nearest minimal window of the rule tail until the maximum allowed delay. If such a rule window is found, then it is added to the sorted list of rule windows maintaining the order. A single call to \nextrwin for a trigger of rule \(r\), in the worst-case, incurs computational time in the order of \(\vert \rtaill{r} \vert^2\) to find the next nearest minimal window of the rule tail (ignoring the parameters \maxgap and \maxdelay). Suppose \(W\) denotes the sorted list of rule windows at any point in time. Once (if) the next best rule window is found, the method incurs additional computational time in the order of \(\;\log_2(\vert W\vert)\) to find the position of insertion using a binary search. 

The question is how many such calls to \nextrwin could happen in the worst case. We could also upper bound the size of \(W\) by the same value. To answer this question, let us consider when \nextrwin is called. It is called whenever an event that participates in a rule window is already covered by a previous rule window. From the point of view of a rule trigger, each event following it until a limit determined by \maxdelay and \maxgap times the rule tail, can potentially participate in a rule window. The \miner starts with one such rule window and looks for the next best rule window if and only if any of the participating events is already covered. Further, the next best rule window omits the previously covered events. Therefore, the maximum number of times \nextrwin gets called is limited by the number of events following the rule trigger, given by \(\vert \rtaill{r} \vert * (\maxdelay + \maxgap + 1)\). Ignoring the constants for the purpose of complexity analysis, we can rewrite it as \(\vert \rtaill{r} \vert\). Over all triggers for all rules, we can then compute the worst-case time complexity of finding the next best windows and adding them to the sorted list of rule windows as 
\[
    \mathcal{O}\left(\sum_{r \in R} \vert S \vert * \vert \rtaill{r}\vert * \left(\vert \rtaill{r} \vert^2 + \log_2(\vert W\vert)\right)\right)
\;,
\]
where \(W\) is the list of rule windows. Once again, as worst-case, we use \varname{max\_tail\_size} to rewrite the same. Further, we can limit the size up to which \(W\) can grow by the number of times \nextrwin gets called, i.e in the order of \(\vert R \vert * \vert S \vert * \varname{max\_tail\_size}\). Putting it all together, we find the total computational time for all calls to \nextrwin to be in the order of  
\begin{equation}
\label{appeqnnbw}
    \mathcal{O}(\vert R \vert * \vert S \vert * \varname{max\_tail\_size} * \log_2(\vert R \vert * \vert S \vert * \varname{max\_tail\_size})+ \vert R \vert * \vert S \vert * \varname{max\_tail\_size}^3)
\end{equation}
Finally, using the ordered list of rule windows \(W\), we cover the sequence \(S\). As singleton rules are also included in \(R\), it is guaranteed to cover the entire sequence in one iteration over all the rule windows in \(W\) (in practice, it will be much lesser as many events get covered by a single rule window). Therefore, we can compute the worst-case time complexity to loop over the list of rule windows and cover the sequence \(S\) as 
\begin{equation}
\label{appeqncvr}
    \mathcal{O}(\vert R\vert * \vert S \vert * \varname{max\_tail\_size})\;.
\end{equation}
Thus, from equations \ref{appeqnfrw}, \ref{appeqnsrt}, \ref{appeqnnbw} and \ref{appeqncvr}, we can compute the total worst-case time complexity of \cover as
\[
\begin{aligned}
    \mathcal{O}(&\vert R \vert * \vert S \vert * (\varname{max\_head\_size} + \varname{max\_tail\_size}^2)\\
    &+
    \vert R\vert * \vert S\vert * \log_2(\vert R\vert * \vert S\vert)\\
    &+
    \vert R\vert * \vert S \vert * \varname{max\_tail\_size}\\
    &+
    \vert R \vert * \vert S \vert * \varname{max\_tail\_size} * \log_2(\vert R \vert * \vert S \vert * \varname{max\_tail\_size})\\
    &+
    \vert R \vert * \vert S \vert * \varname{max\_tail\_size}^3)\;.
\end{aligned}
\]
Considering only the dominating terms, we get 
\begin{equation}
\begin{aligned}
\label{eqngreedycover}
    \mathcal{O}(\vert R \vert * \vert S \vert * (&\varname{max\_head\_size} \\
    &+ \varname{max\_tail\_size}^3 \\
    &+ \varname{max\_tail\_size} * \log_2(\vert R \vert * \vert S \vert * \varname{max\_tail\_size}))\;.
\end{aligned}
\end{equation}

\paragraph{Time Complexity of \ourmethodmine}
Next, we analyze the time complexity of the greedy miner, \ourmethodmine. Let us consider a single iteration of the miner. Let \(R'\) be the candidate rule set at that time point. Then, \ourmethodmine grows the rule set by searching for a new rule that improves the encoding cost by extending each rule, at each position, with their significant neighbors. As worst-case, let us assume that the miner had to search over all rules, at all positions. Further, let us assume that all events in the alphabet $\Omega$ are significant (although this is impossible). To simplify the computations, we use \(\max_{r' \in R'} \vert \rheadd{r'}\vert\) as the \varname{max\_head\_size} and \(\max_{r' \in R'} \vert \rtaill{r'}\vert\) as the \varname{max\_tail\_size} to upper bound the lengths of \rheadd{r'} and \rtaill{r'} for any \(r' \in R'\). Then, the computational time of the search will be in the order of \(
\mathcal{O}(\vert R' \vert * (\varname{max\_head\_size} + \varname{max\_tail\_size}) * \vert \Omega \vert)\;.
\)
Once a new rule is added to the rule set, \ourmethodmine tries to prune the rule set by removing each non-singleton rule. The computational time required by \prune will be in the order of \(\vert R'\vert - \vert \Omega \vert \). Considering only the dominating term, we can thus conclude the worst-case time complexity of each iteration as 
\begin{equation}
    \mathcal{O}(\vert R' \vert * (\varname{max\_head\_size} + \varname{max\_tail\_size}) * \vert \Omega \vert)\;,
\end{equation}
where \(R'\) denotes the candidate rule set at that time point. Next, we try to analyze the number of iterations possible before the algorithm converges.

We know that in each iteration, \ourmethodmine adds a new rule to the current rule set only if the addition improves the encoding cost. Similarly, a rule is pruned from the current rule set only if the removal improves the encoding cost. If the encoding cost cannot be improved anymore, then the algorithm halts. In other words, \ourmethodmine will never revert back to a rule set from which it grew in the past. Therefore, the maximum number of iterations is upper bounded by the number of unique rule sets possible. In Section \ref{ogcomplexity}, we saw that the number of possible rule sets is in the order of \(\mathcal{O}(2^{\vert \Omega \vert^{\vert S \vert}})\). 

\paragraph{Caching for Faster Runtime} In practice, however, we observe the number of rule sets considered by the greedy approach to be much smaller. Further, we cache the rule windows found for each rule as and when they are first encountered. Therefore, if the same rule is present in a future rule set, we do not recompute the rule windows. In other words, \bestrwins is invoked only once per rule. The same is true for \signineigh. We cache the neighbors found for each rule as and when they are first encountered. As a result, the time complexity in practice would be much lower, even if \ourmethodmine attempted the worst-case possibility of all unique rule sets before converging. Further, we do not reconsider rules once pruned in the future iterations.

	\fi

\end{document}